\documentclass[11pt,letterpaper]{article}
\usepackage[letterpaper]{geometry}
\usepackage{acl2012}
\usepackage{times}
\usepackage{latexsym}
\usepackage{amsmath}
\usepackage{multirow}
\usepackage{url}
\makeatletter
\newcommand{\@BIBLABEL}{\@emptybiblabel}
\newcommand{\@emptybiblabel}[1]{}
\makeatother
\usepackage[hidelinks]{hyperref}
\usepackage[font=normalsize]{subfig}
\usepackage[noend]{algpseudocode}

\usepackage{amssymb}
\usepackage{bm}
\usepackage{balance}
\usepackage{arydshln}
\usepackage{mathtools}

\usepackage{xspace}

\newcommand{\eg}{\emph{e.g.,}\xspace}
\newcommand{\ie}{\emph{i.e.,}\xspace}

\title{Context Gates for Neural Machine Translation}

\def\sstaff{$^\ddag$}
\def\fndaff{$^\dagger$}
\author{Zhaopeng Tu\fndaff ~~~ Yang Liu\sstaff ~~~ Zhengdong Lu\fndaff ~~~ Xiaohua Liu\fndaff ~~~ Hang Li\fndaff
\\
\\
{ \fndaff {Noah's Ark Lab, Huawei Technologies, Hong Kong}}   \\
{ \tt \{tu.zhaopeng,lu.zhengdong,liuxiaohua3,hangli.hl\}@huawei.com}\\
{ \sstaff {Department of Computer Science and Technology, Tsinghua University, Beijing}}\\
{ \tt liuyang2011@tsinghua.edu.cn}\\
}

\begin{document}
\maketitle

\begin{abstract}
\noindent
In neural machine translation (NMT), generation of a target word depends on both source and target contexts. We find that source contexts have a direct impact on the {\em adequacy} of a translation while target contexts affect the {\em fluency}. Intuitively, generation of a content word should rely more on the source context and generation of a functional word should rely more on the target context. Due to the lack of effective control over the influence from source and target contexts, conventional NMT tends to yield fluent but inadequate translations. To address this problem, we propose {\em context gates} which dynamically control the ratios at which source and target contexts contribute to the generation of target words. In this way, we can enhance both the adequacy and fluency of NMT with more careful control of the information flow from contexts. Experiments show that our approach significantly improves upon a standard attention-based NMT system by +2.3 BLEU points.
\end{abstract}

\section{Introduction}

\begin{table}[t]
\centering
\subfloat{
\begin{tabular}{c|p{0.75\columnwidth}}
input &  j{\=\i}nni{\' a}n  qi{\' a}n  li{\v a}ng  yu{\` e}  gu{\v a}ngd{\= o}ng  g{\= a}ox{\=\i}n  j{\`\i}sh{\` u}  ch{\v a}np{\v\i}n  ch{\= u}k{\v o}u  37.6y{\`\i}  m{\v e}iyu{\' a}n\\
\hline
NMT & {\em in the first two months of this year ,} the export of new high level technology product was UNK - billion us dollars\\
\hline
$\bigtriangledown src$ & china 's guangdong hi - tech exports hit 58 billion dollars\\
\hline
$\bigtriangledown tgt$ & china 's export of high and new hi - tech exports of {\em the export of the export of the export of the export of the export of the export of the export of the export of $\cdots$}\\
\end{tabular}}
\caption{Source and target contexts are highly correlated to translation adequacy and fluency, respectively.
$\bigtriangledown src$ and $\bigtriangledown tgt$ denote halving the contributions from the {\em source} and {\em target} contexts when generating the translation, respectively.}
\label{table-wrong-cases}
\end{table}

Neural machine translation (NMT)~\cite{Kalchbrenner:2013:EMNLP,Sutskever:2014:NIPS,Bahdanau:2015:ICLR} has made significant progress in the past several years.  Its goal is to construct and utilize a single large neural network to accomplish the entire translation task. One great advantage of NMT is that the translation system can be completely constructed by learning from data without human involvement ({\em cf.,} feature engineering in statistical machine translation (SMT)). The encoder-decoder architecture is widely employed~\cite{Cho:2014:EMNLP,Sutskever:2014:NIPS}, in which the encoder summarizes the source sentence into a vector representation, and the decoder generates the target sentence word-by-word from the vector representation. The representation of the source sentence and the representation of the partially generated target sentence (translation) at each position are referred to as source context and target context, respectively. The generation of a target word is determined jointly by the source context and target context.

Several techniques in NMT have proven to be very effective, including gating~\cite{Hochreite:1997,Cho:2014:EMNLP} and attention~\cite{Bahdanau:2015:ICLR} which can model long-distance dependencies and complicated alignment relations in the translation process. Using an encoder-decoder framework that incorporates gating and attention techniques, it has been reported that the performance of NMT can surpass the performance of traditional SMT as measured by BLEU score~\cite{Luong:2015:EMNLP}.

Despite this success, we observe that NMT usually yields fluent but inadequate translations.\footnote{Fluency measures whether the translation is fluent, while adequacy measures whether the translation is faithful to the original sentence~\cite{Snover:2009:WMT}.}
We attribute this to a stronger influence of target context on generation, which results from a stronger language model than that used in SMT. One question naturally arises: {\em what will happen if we change the ratio of influences from the source or target contexts? }

Table~\ref{table-wrong-cases} shows an example in which an attention-based NMT system~\cite{Bahdanau:2015:ICLR} generates a fluent yet inadequate translation (\eg missing the translation of ``{\em gu{\v a}ngd{\= o}ng}'').
When we halve the contribution from the source context, the result further loses its adequacy by missing the partial translation ``{\em in the first two months of this year}''.
One possible explanation is that the target context takes a higher weight and thus the system favors a shorter translation.
In contrast, when we halve the contribution from the target context, the result completely loses its fluency by repeatedly generating the translation of ``{\em ch{\= u}k{\v o}u}'' (\ie ``{\em the export of}'') until the generated translation reaches the maximum length.
Therefore, this example indicates that {\em source and target contexts in NMT are highly correlated to translation adequacy and fluency, respectively.}

In fact, conventional NMT lacks effective control on the influence of source and target contexts. At each decoding step, NMT treats the source and target contexts equally, and thus ignores the different needs of the contexts. For example, content words in the target sentence are more related to the translation adequacy, and thus should depend more on the source context. In contrast, function words in the target sentence are often more related to the translation fluency (\eg ``{\em of}'' after ``{\em is fond}''), and thus should depend more on the target context.

In this work, we propose to use {\em context gates} to control the contributions of source and target contexts on the generation of target words (decoding) in NMT.
Context gates are non-linear gating units which can dynamically select the amount of context information in the decoding process. Specifically, 
at each decoding step, the context gate examines both the source and target contexts, and outputs a ratio between zero and one to determine the percentages of information to utilize from the two contexts. In this way, the system can balance the adequacy and fluency of the translation with regard to the generation of a word at each position.

Experimental results show that introducing context gates leads to an average improvement of +2.3 BLEU points over a standard attention-based NMT system~\cite{Bahdanau:2015:ICLR}.
An interesting finding is that we can replace the GRU units in the decoder with conventional RNN units and in the meantime utilize context gates. The translation performance is comparable with the standard NMT system with GRU, but the system enjoys a simpler structure (\ie uses only a single gate and half of the parameters) and a faster decoding (\ie requires only half the matrix computations for decoding).\footnote{Our code is publicly available at \protect\url{https://github.com/tuzhaopeng/NMT}.}

\section{Neural Machine Translation}
\label{sec-background}

\begin{figure}[t]
\centering
\includegraphics[width=0.45\textwidth]{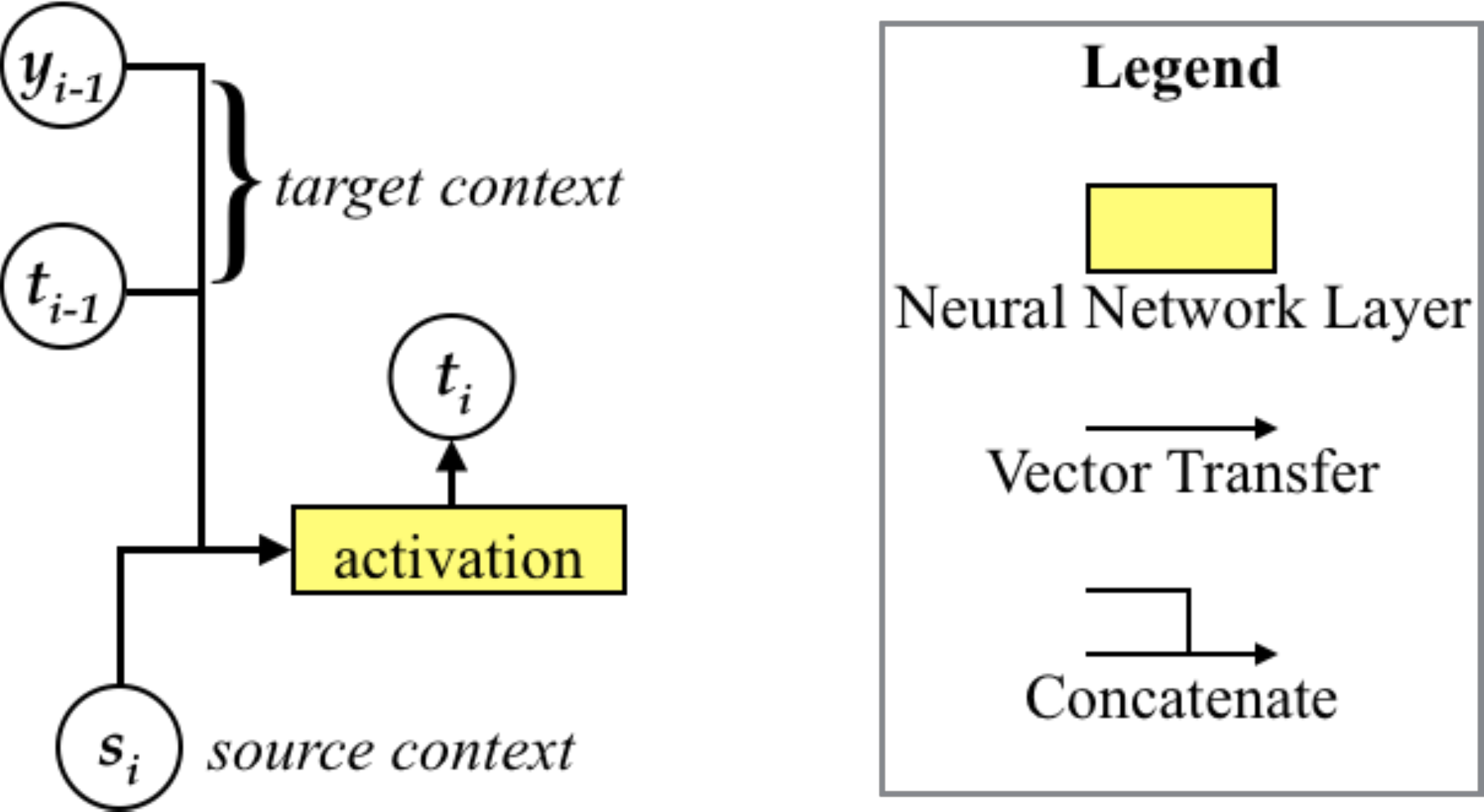}
\caption{Architecture of decoder RNN.}
\label{figure-gated-attention}
\end{figure}

Suppose that ${\bf x}=x_1, \dots x_j, \dots x_J$ represents a source sentence and ${\bf y}=y_1, \dots y_i, \dots y_{I}$ a target sentence. NMT directly models the probability of translation from the source sentence to the target sentence word by word:
\begin{equation}
P({\bf y}|{\bf x}) = \prod_{i=1}^{I} P(y_i| y_{<i}, {\bf x})
\end{equation}
where $y_{<i}=y_1,\dots, y_{i-1}$.
As shown in Figure~\ref{figure-gated-attention}, the probability of generating the {\em i}-th word $y_i$ is computed by using a recurrent neural network (RNN) in the decoder:
\begin{equation}
P(y_i|y_{<i}, {\bf x}) = g(y_{i-1}, t_i, s_i)
\end{equation}
where $g(\cdot)$ first linearly transforms its input then applies a softmax function, $y_{i-1}$ is the previously generated word, $t_i$ is the $i$-th decoding hidden state, and $s_i$ is the $i$-th source representation.
The state $t_i$ is computed as follows:
\begin{eqnarray}
t_i &=& f(y_{i-1}, t_{i-1}, s_i) \nonumber \\
      &=& f(We(y_{i-1}) + Ut_{i-1} + Cs_i)
\label{eqn-hidden-state}
\end{eqnarray}
where
\begin{itemize}
\item $f(\cdot)$ is a function to compute the current decoding state given all the related inputs. It can be either a vanilla RNN unit using $\tanh$ function, or a sophisticated gated RNN unit such as GRU~\cite{Cho:2014:EMNLP} or LSTM~\cite{Hochreite:1997}.
\item $e(y_{i-1}) \in \mathbb{R}^{m}$ is an $m$-dimensional embedding of the previously generated word $y_{i-1}$.
 \item $s_i$ is a vector representation extracted from the source sentence by the encoder. The encoder usually uses an RNN to encode the source sentence ${\bf x}$ into a sequence of hidden states ${\bf h}=h_1, \dots h_j, \dots h_J$, in which $h_j$ is the hidden state of the $j$-th source word $x_j$. $s_i$ can be either a static vector that summarizes the whole sentence (\eg $s_i \equiv h_J$)~\cite{Cho:2014:EMNLP,Sutskever:2014:NIPS}, or a dynamic vector that selectively summarizes certain parts of the source sentence at each decoding step (\eg $s_i = \sum_{j=1}^{J} \alpha_{i,j} h_j$ in which $\alpha_{i,j}$ is alignment probability calculated by an attention model)~\cite{Bahdanau:2015:ICLR}.
 \item $W \in \mathbb{R}^{n \times m}$, $U \in \mathbb{R}^{n \times n}$, $C \in \mathbb{R}^{n \times n'}$ are matrices with $n$ and $n'$ being the numbers of units of decoder hidden state and source representation, respectively.
\end{itemize}

The inputs to the decoder (\ie $s_{i}$, $t_{i-1}$, and $y_{i-1}$) represent the contexts. Specifically, the source representation $s_i$ stands for {\bf source context}, which embeds the information from the source sentence. The previous decoding state $t_{i-1}$ and the previously generated word $y_{i-1}$ constitute the {\bf target context}.\footnote{In a recent implementation of NMT (\protect\url{https://github.com/nyu-dl/dl4mt-tutorial}), $t_{i-1}$ and $y_{i-1}$ are combined together with a GRU before being fed into the decoder, which can boost translation performance. We follow the practice and treat both of them as target context.}

\begin{figure}[t]
\centering
\subfloat[Lengths of translations in words.]{
\includegraphics[width=0.38\textwidth]{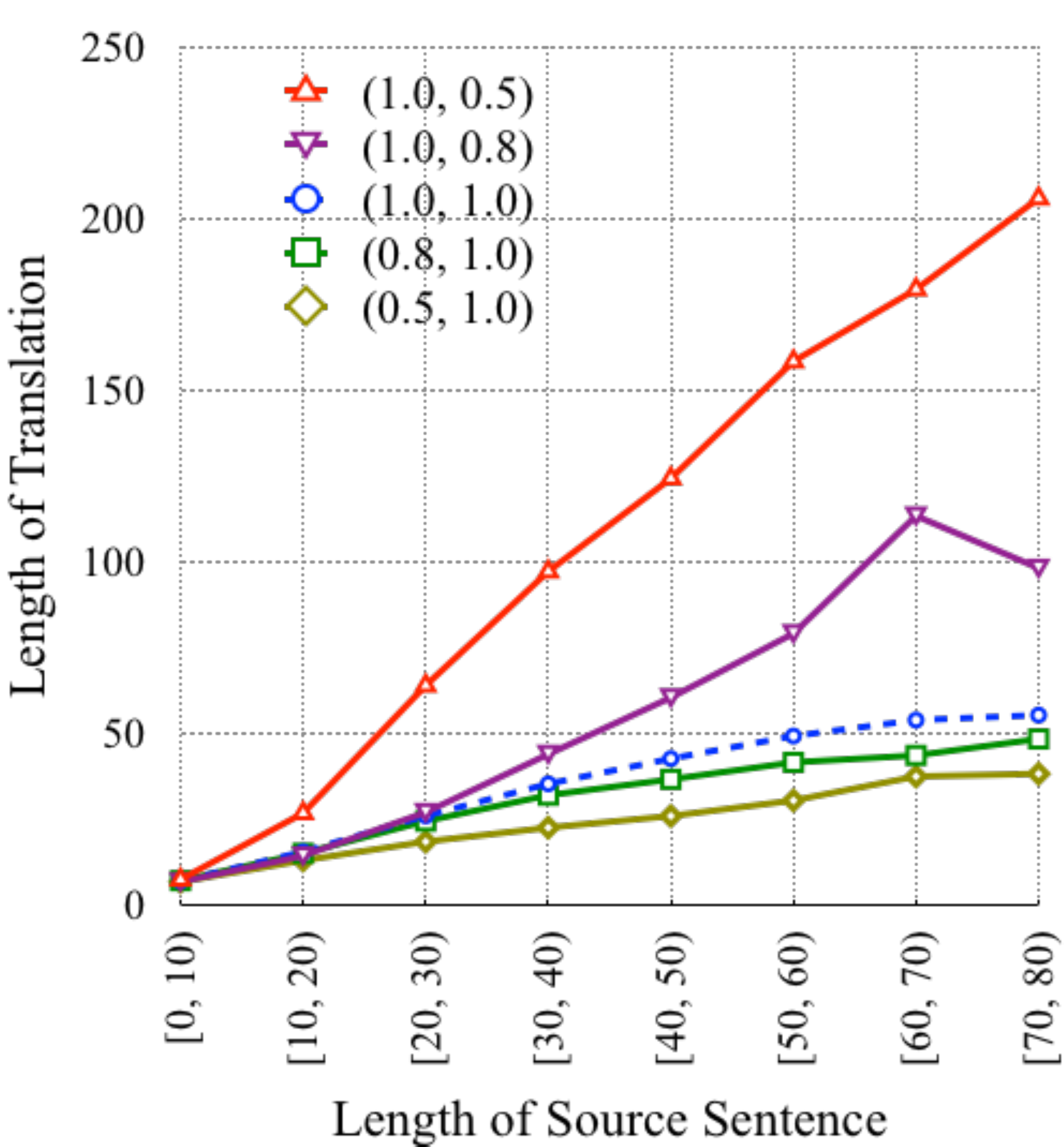}} \\
\subfloat[Subjective evaluation.]{
\includegraphics[width=0.36\textwidth]{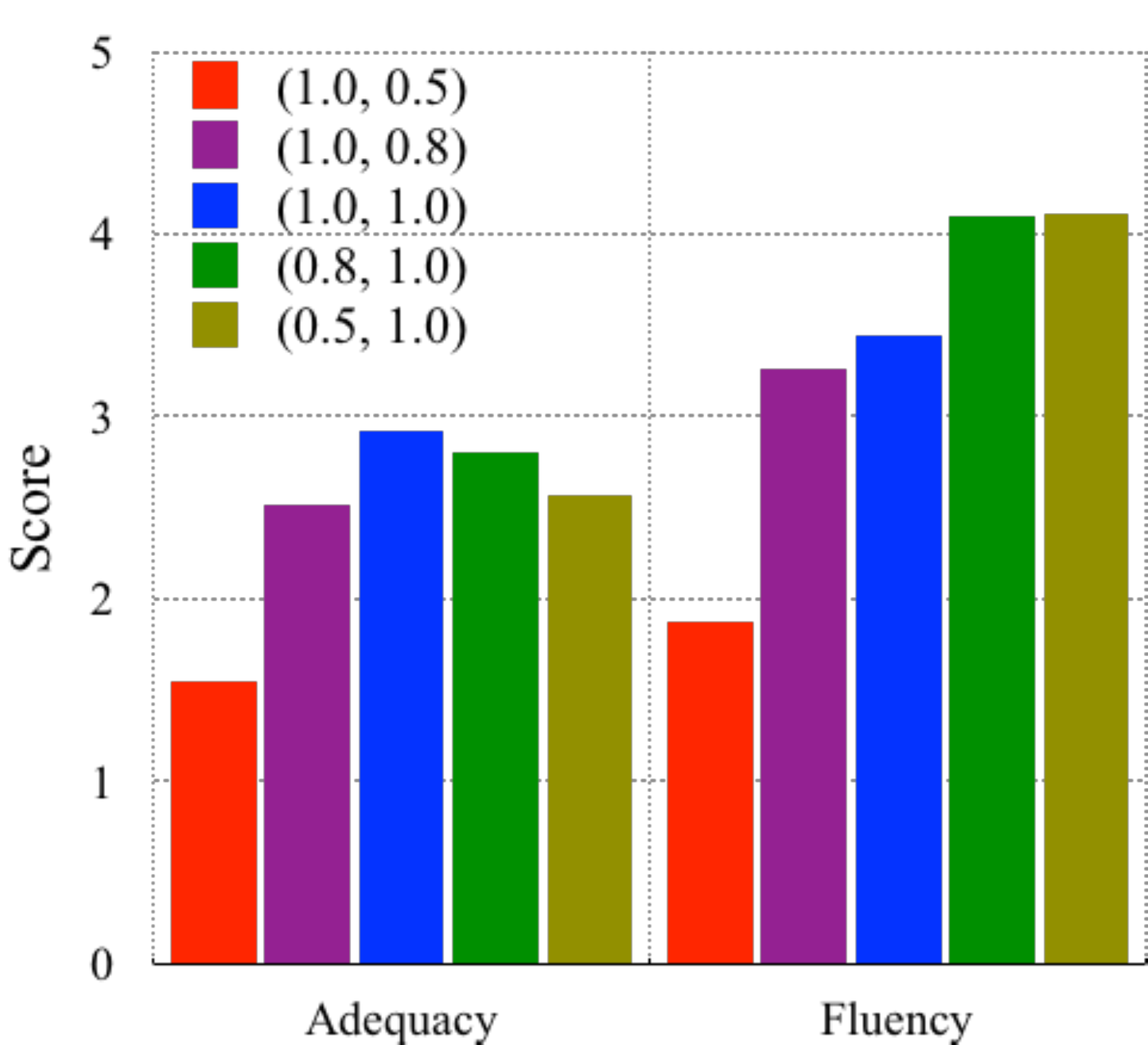}}
\caption{Effects of source and target contexts. The pair $(a, b)$ in the legends denotes scaling source and target contexts with ratios $a$ and $b$ respectively.}
\label{figure-source-target-contexts}
\end{figure}

\subsection{Effects of Source and Target Contexts}
\label{sec-effects-of-contexts}

We first empirically investigate our hypothesis: whether source and target contexts correlate to translation adequacy and fluency.
Figure~\ref{figure-source-target-contexts}(a) shows the translation lengths with various scaling ratios $(a, b)$ for source and target contexts:
\begin{equation}
t_i = f(b \otimes (W_e(y_{i-1}) + U t_{i-1}) + a \otimes C s_i) \nonumber
\end{equation}
For example, the pair (1.0, 0.5) means fully leveraging the effect of source context while halving the effect of target context.
Reducing the effect of target context (\ie the lines (1.0, 0.8) and (1.0, 0.5)) results in longer translations, while reducing the effect of source context (\ie the lines (0.8, 1.0) and (0.5, 1.0)) leads to shorter translations.
When halving the effect of the target context, most of the generated translations reach the maximum length, which is three times the length of source sentence in this work.

Figure~\ref{figure-source-target-contexts}(b) shows the results of manual evaluation on 200 source sentences randomly sampled from the test sets. 
Reducing the effect of source context (\ie (0.8, 1.0) and (0.5, 1.0)) leads to more fluent yet less adequate translations.
On the other hand, reducing the effect of target context (\ie (1.0, 0.5) and (1.0, 0.8)) is expected to yield more adequate but less fluent translations.
In this setting, the source words are translated (\ie higher adequacy) while the translations are in wrong order (\ie lower fluency).
In practice, however, we observe the side effect that some source words are translated repeatedly until the translation reaches the maximum length (\ie lower fluency), while others are left untranslated (\ie lower adequacy).
The reason is two fold:
\begin{enumerate}
  \item NMT lacks a mechanism that guarantees that each source word is translated.\footnote{The recently proposed coverage based technique can alleviate this problem~\cite{Tu:2016:ACL}. In this work, we consider another approach, which is complementary to the coverage mechanism.} The decoding state implicitly models the notion of ``coverage'' by recurrently reading the time-dependent source context $s_i$. Lowering its contribution weakens the ``coverage'' effect and encourages the decoder to regenerate phrases multiple times to achieve the desired translation length.

  \item The translation is incomplete. As shown in Table~\ref{table-wrong-cases}, NMT can get stuck in an infinite loop repeatedly generating a phrase due to the overwhelming influence of the source context. As a result, generation terminates early because the translation reaches the maximum length allowed by the implementation, even though the decoding procedure is not finished. 
\end{enumerate}

The quantitative (Figure~\ref{figure-source-target-contexts}) and qualitative (Table~\ref{table-wrong-cases}) results confirm our hypothesis, \ie source and target contexts are highly correlated to translation adequacy and fluency.
We believe that a mechanism that can dynamically select information from source context and target context would be useful for NMT models, and this is exactly the approach we propose.

\section{Context Gates}
\label{sec-gated-attention}

\subsection{Architecture}

Inspired by the success of gated units in RNN~\cite{Hochreite:1997,Cho:2014:EMNLP}, we propose using {\em context gates} to dynamically control the amount of information flowing from the source and target contexts and thus balance the fluency and adequacy of NMT at each decoding step.

Intuitively, at each decoding step $i$, the context gate looks at input signals from both the source (\ie $s_i$) and target (\ie $t_{i-1}$ and $y_{i-1}$) sides, and outputs a number between $0$ and $1$ for each element in the input vectors, where $1$ denotes ``completely transferring this'' while $0$ denotes ``completely ignoring this''. The corresponding input signals are then processed with an element-wise multiplication before being fed to the activation layer to update the decoding state.

\begin{figure}[t]
\centering
\includegraphics[width=0.26\textwidth]{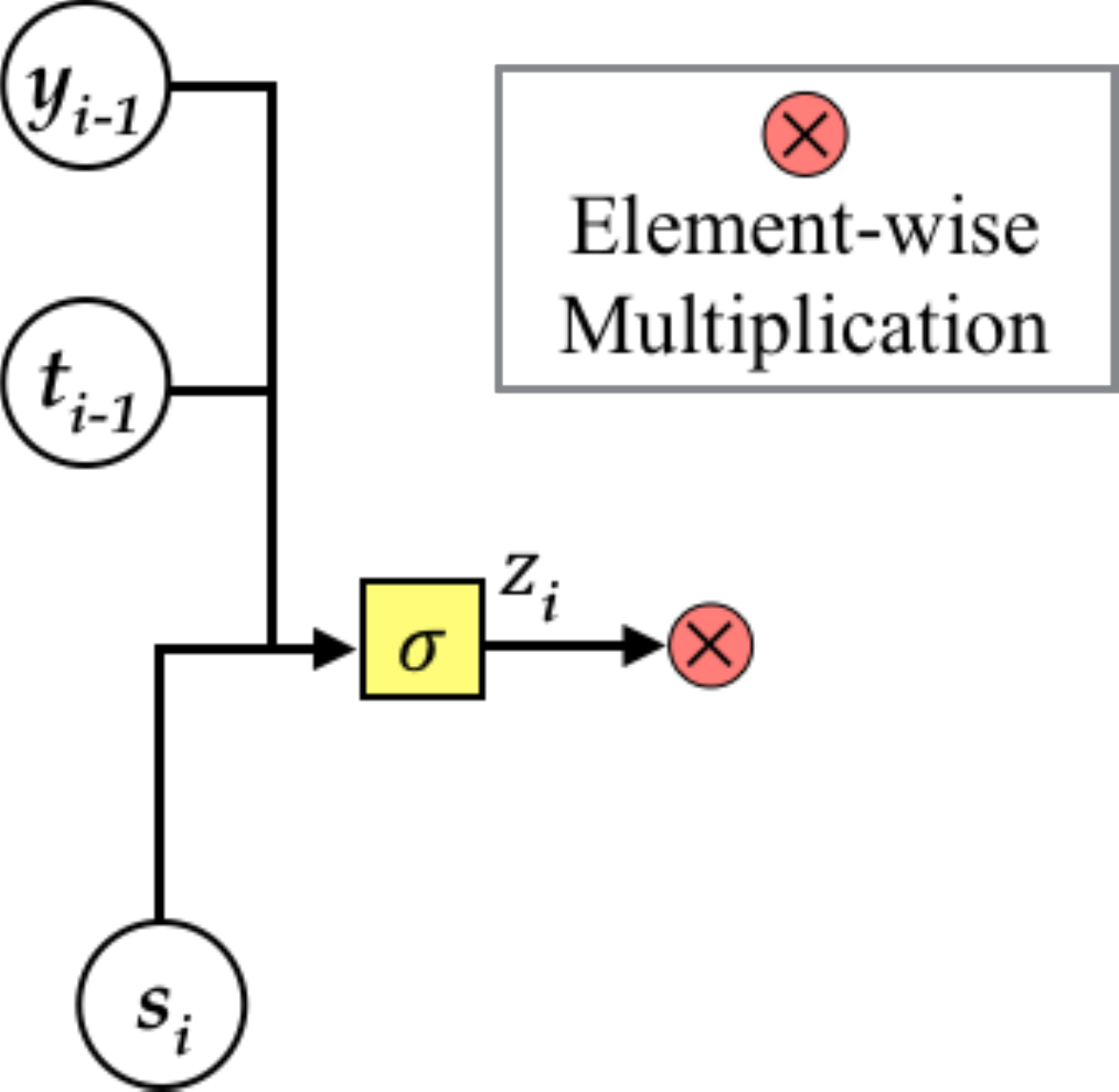}
\caption{Architecture of context gate. }
\label{figure-architecture}
\end{figure}

\begin{figure*}[t]
\centering
\subfloat[Context Gate ({\em source})]{
\includegraphics[width=0.26\textwidth]{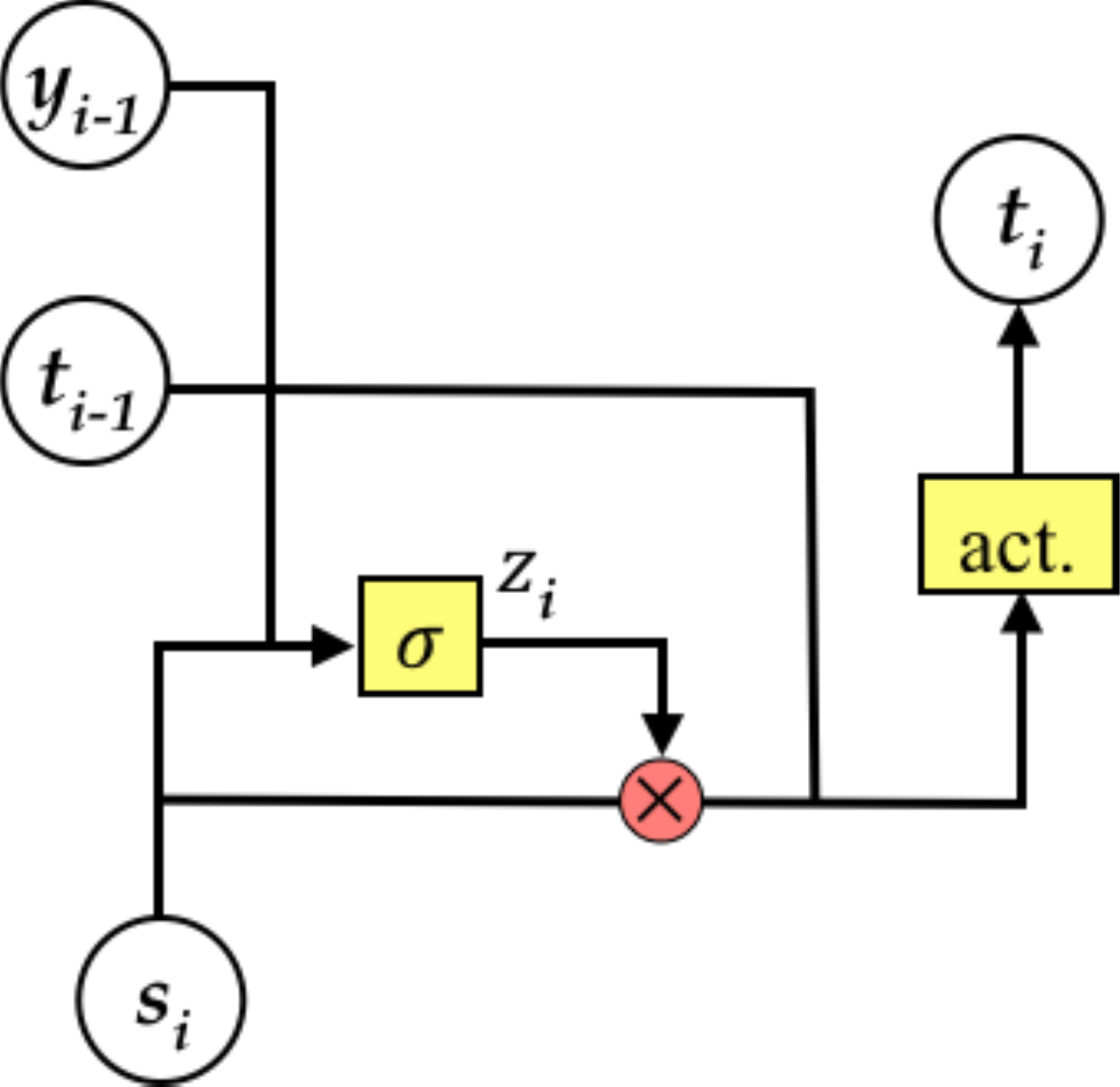}
}
\hfill
\subfloat[Context Gate ({\em target})]{
\includegraphics[width=0.26\textwidth]{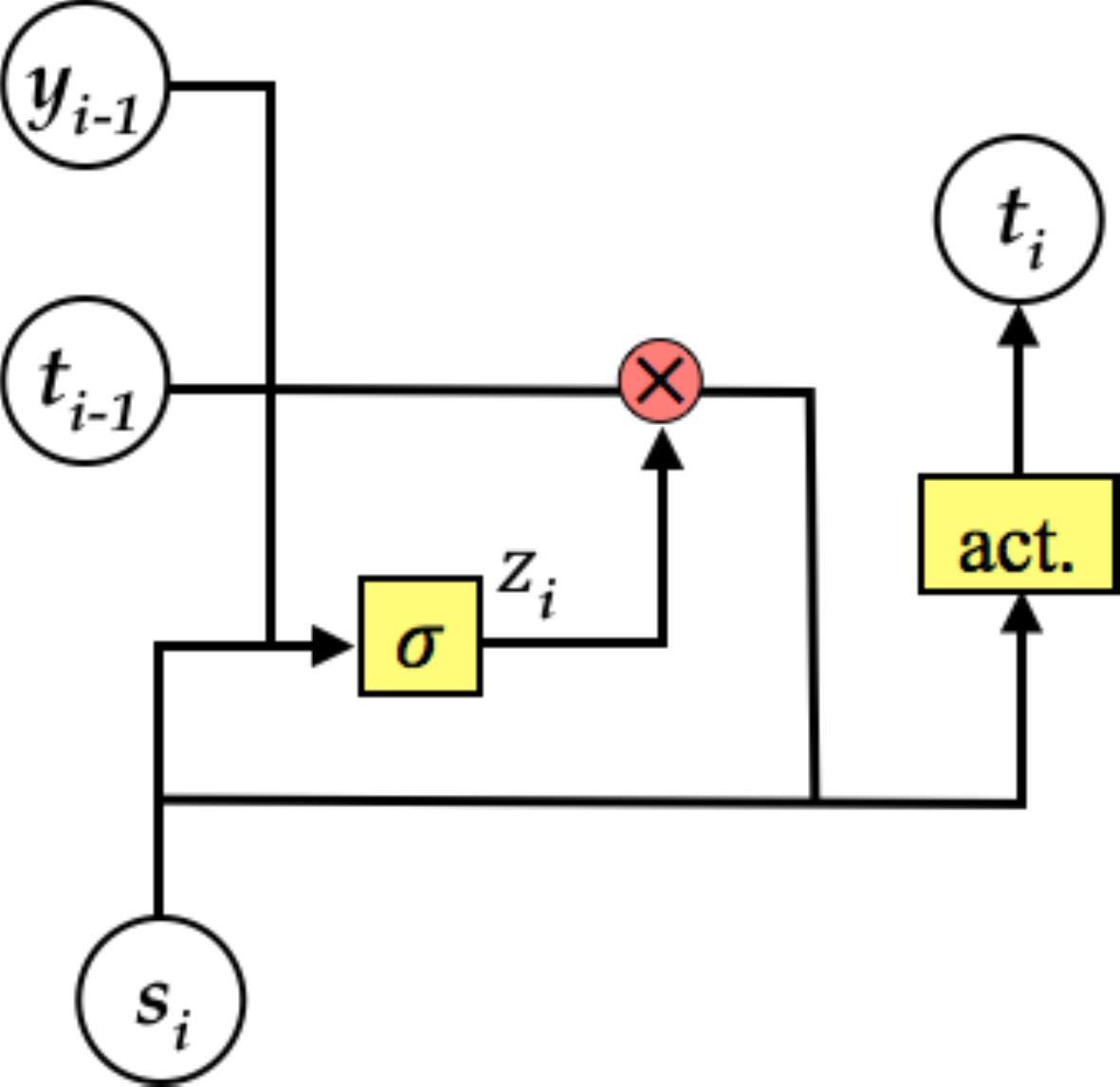}
}
\hfill
\subfloat[Context Gate ({\em both})]{
\includegraphics[width=0.26\textwidth]{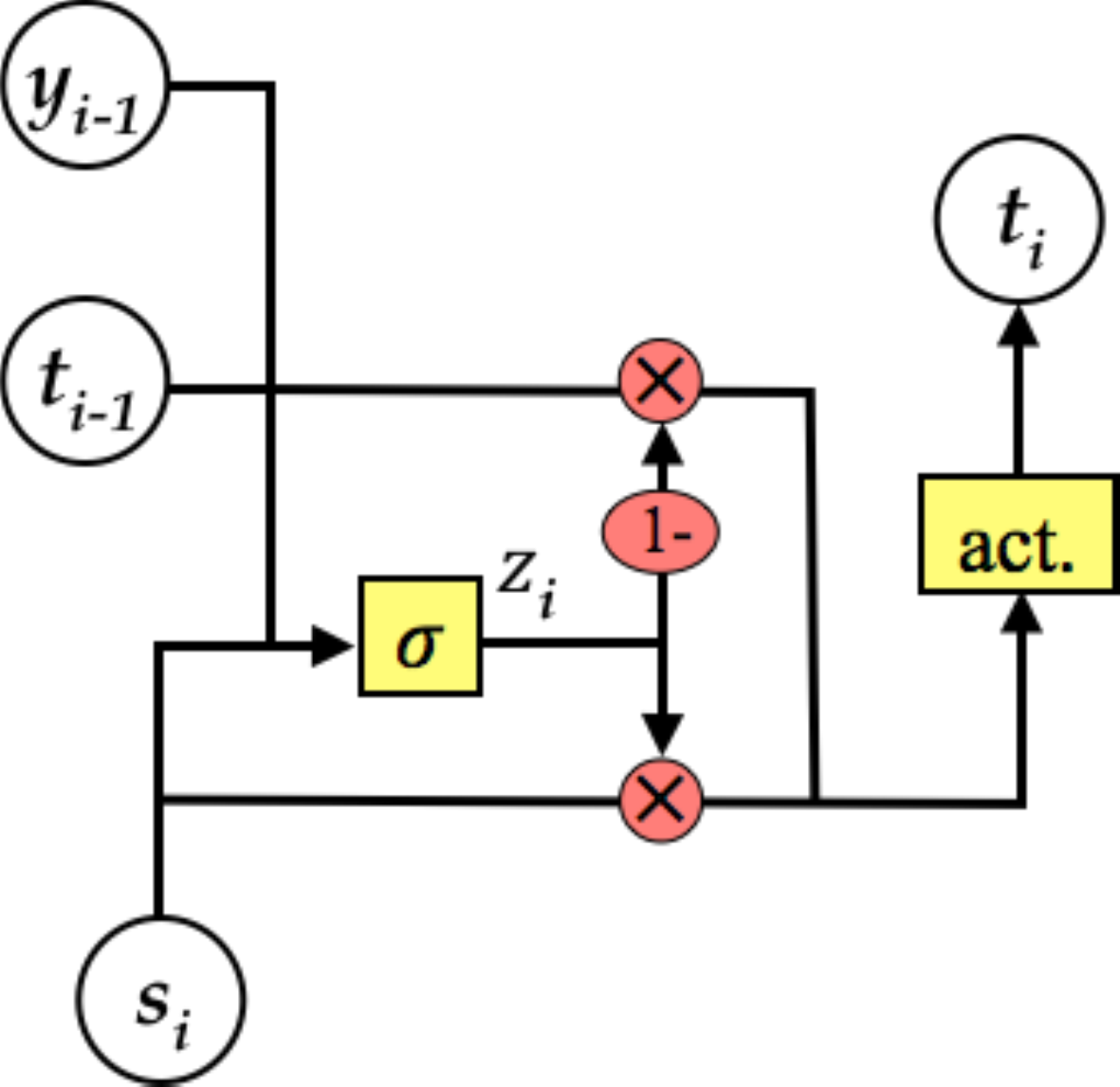}
}
\caption{Architectures of NMT with various context gates, which either scale only one side of translation contexts (\ie source context in (a) and target context in (b)) or control the effects of both sides (\ie (c)).}
\label{figure-context-gates}
\end{figure*}

Formally, a context gate consists of a sigmoid neural network layer and an element-wise multiplication operation, as illustrated in Figure~\ref{figure-architecture}. The context gate assigns an element-wise weight to the input signals, computed by
\begin{equation}
z_i =  \sigma (W_z e(y_{i-1}) + U_z t_{i-1} + C_z s_i)
\label{eqn-context-gate}
\end{equation}
Here $\sigma(\cdot)$ is a logistic sigmoid function, and $W_z \in \mathbb{R}^{n \times m}$, $U_z \in \mathbb{R}^{n \times n}$, $C_z \in \mathbb{R}^{n \times n'}$ are the weight matrices. Again, $m$, $n$ and $n'$ are the dimensions of word embedding, decoding state, and source representation, respectively.
Note that $z_i$ has the same dimensionality as the transferred input signals (\eg $Cs_i$), and thus each element in the input vectors has its own weight.

\subsection{Integrating Context Gates into NMT}

Next, we consider how to integrate context gates into an NMT model.

The context gate can decide the amount of context information used in generating the next target word at each step of decoding.
For example, after obtaining the partial translation ``{\em \dots new high level technology product}'', the gate looks at the translation contexts and decides to depend more heavily on the source context. Accordingly, the gate assigns higher weights to the source context and lower weights to the target context and then feeds them into the decoding activation layer. This could correct inadequate translations, such as the missing translation of ``{\em gu{\v a}ngd{\= o}ng}'', due to greater influence from the target context.

We have three strategies for integrating context gates into NMT that either affect one of the translation contexts or both contexts, as illustrated in Figure~\ref{figure-context-gates}. The first two strategies are inspired by output gates in LSTMs~\cite{Hochreite:1997}, which control the amount of memory content utilized. In these kinds of models, $z_i$ only affects either source context (\ie $s_i$) or target context (\ie $y_{i-1}$ and $t_{i-1}$):
\begin{itemize}
   \item {\bf Context Gate ({\em source})}
\begin{flalign}
t_i =  f\big(~We(y_{i-1}) + Ut_{i-1} + z_i \circ C s_i ~\big) \nonumber
\end{flalign}
   \item {\bf Context Gate ({\em target})}
\begin{flalign}
t_i =  f\big(~z_i \circ (We(y_{i-1}) + Ut_{i-1}) + C s_i ~\big)  \nonumber
\end{flalign}
\end{itemize}
where $\circ$ is an element-wise multiplication, and $z_i$ is the context gate calculated by Equation~\ref{eqn-context-gate}.
This is also essentially similar to the {\em reset gate} in the GRU, which decides what information to forget from the previous decoding state before transferring that information to the decoding activation layer. The difference is that here the ``reset'' gate resets the context vector rather than the previous decoding state.

The last strategy is inspired by the concept of {\em update gate} from GRU, which takes a linear sum between the previous state $t_{i-1}$ and the candidate new state $\tilde{t}_i$. In our case, we take a linear interpolation between source and target contexts:
\begin{itemize}
   \item {\bf Context Gate ({\em both})}
\begin{flalign}
t_i =  f\big(~&(1-z_i) \circ (We(y_{i-1}) + Ut_{i-1}) \nonumber \\
                                 &+ z_i \circ Cs_i~\big) \nonumber
\end{flalign}
\end{itemize}

\section{Related Work}

\paragraph{Comparison to (Xu et al., 2015):} Context gates are inspired by the gating scalar model proposed by~\newcite{Xu:2015:ICML} for the image caption generation task. The essential difference lies in the task requirement:
\begin{itemize}
\item In image caption generation, the source side (\ie image) contains more information than the target side (\ie caption). Therefore, they employ a gating scalar to scale only the source context.
\item In machine translation, both languages should contain equivalent information. Our model jointly controls the contributions from the source and target contexts. A direct interaction between input signals from both sides is useful for balancing adequacy and fluency of NMT.
\end{itemize}

\begin{figure}[t]
\centering
\subfloat[Gating Scalar]{
\includegraphics[width=0.18\textwidth]{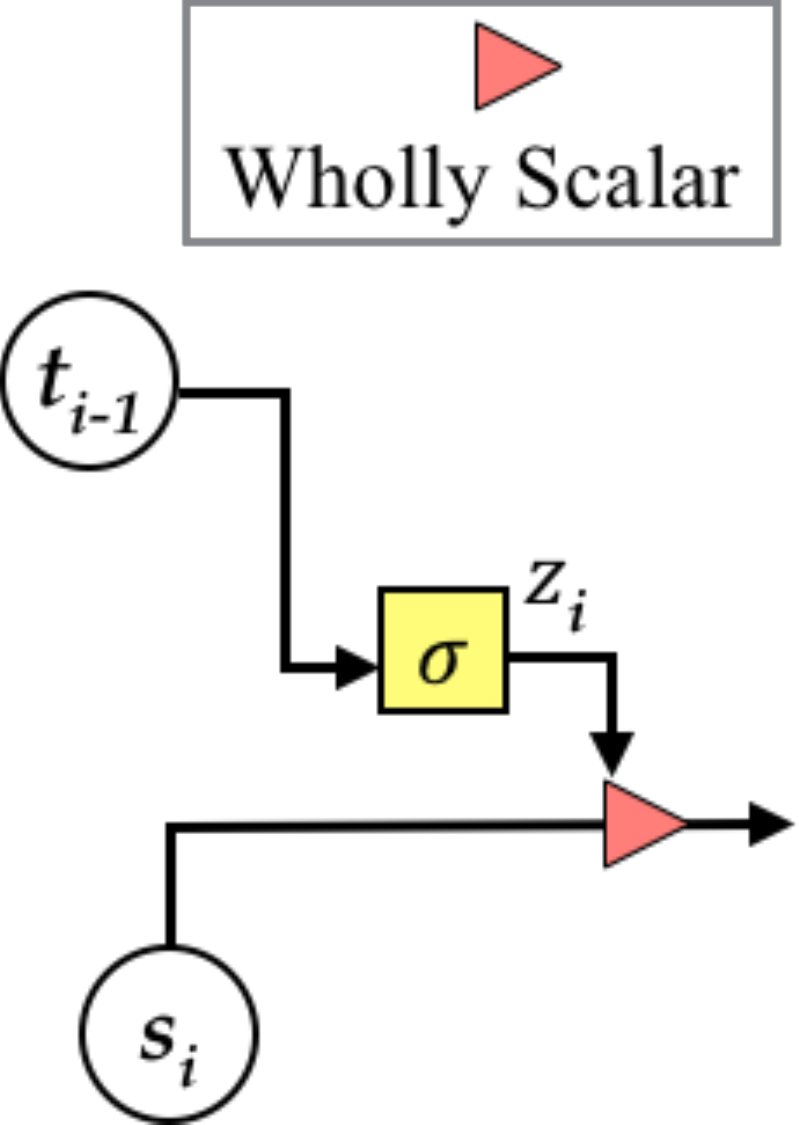}
}
\hfill
\subfloat[Context Gate]{
\includegraphics[width=0.18\textwidth]{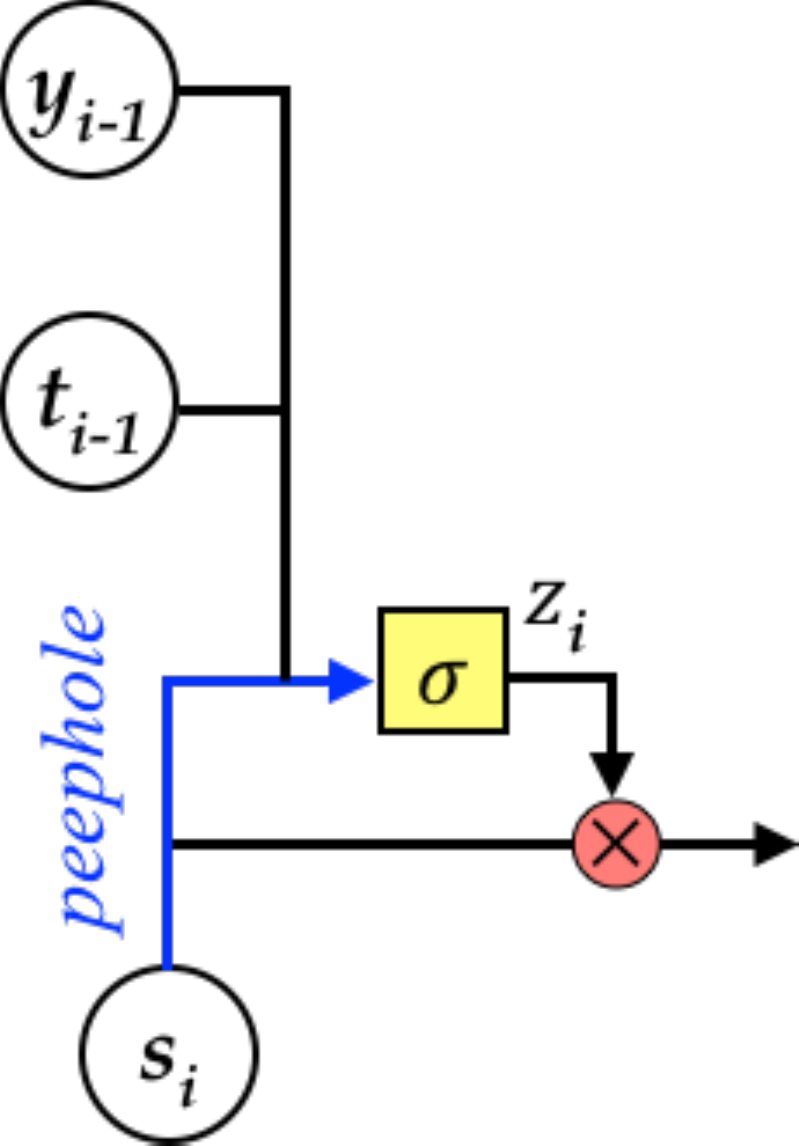}
}
\caption{Comparison to Gating Scalar proposed by~\protect\newcite{Xu:2015:ICML}.} 
\label{figure-comparison}
\end{figure}

Other differences in the architecture include:
\begin{itemize}
\item [1]~\newcite{Xu:2015:ICML} uses a scalar that is shared by all elements in the source context, while we employ a gate with a distinct weight for each element. The latter offers the gate a more precise control of the context vector, since different elements retain different information.
\item [2] We add peephole connections to the architecture, by which the source context controls the gate. It has been shown that peephole connections make precise timings easier to learn~\cite{Gers:2000:IJCNN}.
\item [3]
Our context gate also considers the previously generated word $y_{i-1}$ as input. The most recently generated word can help the gate to better estimate the importance of target context, especially for the generation of function words in translations that may not have a corresponding word in the source sentence (\eg ``of'' after ``is fond'').
\end{itemize}
Experimental results (Section~\ref{sec-component-analysis}) show that these modifications consistently improve translation quality.

\paragraph{Comparison to Gated RNN:} State-of-the-art NMT models~\cite{Sutskever:2014:NIPS,Bahdanau:2015:ICLR} generally employ a gated unit (\eg GRU or LSTM) as the activation function in the decoder.
One might suspect that the context gate proposed in this work is somewhat redundant, given the existing gates that control the amount of information carried over from the previous decoding state $s_{i-1}$ (\eg reset gate in GRU). We argue that they are in fact complementary:
the context gate regulates the contextual information flowing into the decoding state, while the gated unit captures long-term dependencies between decoding states.
Our experiments confirm the correctness of our hypothesis: the context gate not only improves translation quality when compared to a conventional RNN unit (\eg an element-wise $tanh$), but also when compared to a gated unit of GRU, as shown in Section~\ref{sec-results}.

\paragraph{Comparison to Coverage Mechanism:} Recently,~\newcite{Tu:2016:ACL} propose adding a coverage mechanism into NMT to alleviate over-translation and under-translation problems, which directly affect translation adequacy. They maintain a coverage vector to keep track of which source words have been translated. The coverage vector is fed to the attention model to help adjust future attention. This guides NMT to focus on the un-translated source words while avoiding repetition of source content.
Our approach is complementary: the coverage mechanism produces a better source context representation, while our context gate controls the effect of the source context based on its relative importance.
Experiments in Section~\ref{sec-results} show that combining the two methods can further improve translation performance.
There is another difference as well: the coverage mechanism is only applicable to attention-based NMT models, while the context gate is applicable to all NMT models.

\paragraph{Comparison to Exploiting Auxiliary Contexts in Language Modeling:} A thread of work in language modeling (LM) attempts to exploit auxiliary sentence-level or document-level context in an RNN LM~\cite{Mikolov:2012:SLT,Ji:2015:ICLR,Wang:2016:ACL}.
Independent of our work,~\newcite{Wang:2016:ACL} propose ``early fusion'' models of RNNs where additional information from an inter-sentence context is ``fused'' with the input to the RNN.
Closely related to~\newcite{Wang:2016:ACL}, our approach aims to dynamically control the contributions of required source and target contexts for machine translation, while theirs focuses on integrating auxiliary corpus-level contexts for language modelling to better approximate the corpus-level probability. In addition, we employ a gating mechanism to produce a dynamic weight at different decoding steps to combine source and target contexts, while they do a linear combination of intra-sentence and inter-sentence contexts with static weights. Experiments in Section~\ref{sec-results} show that our gating mechanism significantly outperforms linear interpolation when combining contexts.

\paragraph{Comparison to Handling Null-Generated Words in SMT:} In machine translation, there are certain syntactic elements of the target language that are missing in the source (\ie~{\em null-generated words}). In fact this was the preliminary motivation for our approach: current attention models lack a mechanism to control the generation of words that do not have a strong correspondence on the source side. 
The model structure of NMT is quite similar to the traditional word-based SMT~\cite{Brown:1993:CL}. 
Therefore, techniques that have proven effective in SMT may also be applicable to NMT.
\newcite{Toutanova:2002:EMNLP} extend the calculation of translation probabilities to include null-generated target words in word-based SMT. These words are generated based on both the special source token {\em null} and the neighbouring word in the target language by a mixture model.
We have simplified and generalized their approach: we use context gates to dynamically control the contribution of source context. When producing null-generated words, the context gate can assign lower weights to the source context, by which the source-side information have less influence. In a sense, the context gate relieves the need for a {\em null} state in attention.

\section{Experiments}
\label{sec-experiments}

\begin{table*}[t]
\centering
\begin{tabular}{c|l|c|llll}
    \#	&	{\bf System}	&	{\bf \#Parameters}	&	{\bf MT05}  &  {\bf MT06}	  &	{\bf MT08}  &  {\bf Ave.}	\\
    \hline
    1	&	Moses	              	   		&	--	&	31.37	&	30.85	&	23.01	&	28.41\\
    \hline
    \hline
    2	&	GroundHog ($vanilla$)				&	77.1M	&	26.07	&	27.34	&	20.38	&	24.60\\
    3	&	2 + Context Gate ($both$)      		&	80.7M	&	30.86$^*$	&	30.85$^*$	&	24.71$^*$	&	28.81	\\
    \hline
    \hline
    4	&	GroundHog ($GRU$)		   			&	84.3M	&	30.61	&	31.12	&	23.23	&	28.32	\\
    5	&	4 + Context Gate ($source$)              	&	87.9M	&	31.96$^*$	&	32.29$^*$	&	24.97$^*$	&	29.74	\\
    6	&	4 + Context Gate ($target$)              	&	87.9M	&	32.38$^*$	&	32.11$^*$	&	23.78	&	29.42	\\
    7	&	4 + Context Gate ($both$)              	&	87.9M	&	33.52$^*$	&	33.46$^*$	&	24.85$^*$	&	30.61	\\
    \hline
    \hline
    8	&	GroundHog-Coverage ($GRU$)		   	&	84.4M	&	32.73	&	32.47	&	25.23	&	30.14	\\
    9	&	8 + Context Gate ($both$)              	&	88.0M	&	{\bf 34.13}$^*$	&	{\bf 34.83}$^*$	&	{\bf 26.22}$^*$	&	{\bf 31.73}	\\
\end{tabular}
\caption{Evaluation of translation quality measured by case-insensitive BLEU score. ``GroundHog ($vanilla$)'' and ``GroundHog ($GRU$)'' denote attention-based NMT (Bahdanau et al.,2015) and uses a simple $\tanh$ function or a sophisticated gate function $GRU$ respectively as the activation function in the decoder RNN. ``GroundHog-Coverage'' denotes attention-based NMT with a coverage mechanism to indicate whether a source word is translated or not (Tu et al., 2016). ``*'' indicate statistically significant difference ($p < 0.01$) from the corresponding NMT variant. ``2 + Context Gate ($both$)'' denotes integrating ``Context Gate ($both$)'' into the baseline system in Row 2 (\ie ``GroundHog ($vanilla$)'').}
\label{table-translation-results}
\end{table*}

\subsection{Setup}

We carried out experiments on Chinese-English translation.
The training dataset consisted of 1.25M sentence pairs extracted from LDC corpora\footnote{The corpora include LDC2002E18, LDC2003E07, LDC2003E14, Hansards portion of LDC2004T07, LDC2004T08 and LDC2005T06.}, with 27.9M Chinese words and 34.5M English words respectively.
We chose the NIST 2002 (MT02) dataset as the development set, and the NIST 2005 (MT05), 2006 (MT06) and 2008 (MT08) datasets as the test sets.
We used the case-insensitive 4-gram NIST BLEU score~\cite{Papineni:2002} 
as the evaluation metric, and \emph{sign-test}~\cite{Collins:2005} for the statistical significance test.

For efficient training of the neural networks, we limited the source and target vocabularies to the most frequent 30K words in Chinese and English, covering approximately 97.7\% and 99.3\% of the data in the two languages respectively.
All out-of-vocabulary words were mapped to a special token \texttt{\small UNK}.
We trained each model on sentences of length up to 80 words in the training data. The word embedding dimension was 620 and the size of a hidden layer was 1000.
We trained our models until the BLEU score on the development set stops improving.

We compared our method with representative SMT and NMT\footnote{There is some recent progress on aggregating multiple models or enlarging the vocabulary(\eg, in ~\cite{Jean:2015:ACL}), but here we focus on the generic models.} models:
\begin{itemize}
    \item {\bf Moses}~\cite{Koehn:2007:ACL}: an open source phrase-based translation system with default configuration and a 4-gram language model trained on the target portion of training data;
    \item {\bf GroundHog}~\cite{Bahdanau:2015:ICLR}:  an open source attention-based NMT model with default setting. We have two variants that differ in the activation function used in the decoder RNN:
    1) {\em GroundHog ($vanilla$)} uses a simple $\tanh$ function as the activation function, and 2) {\em GroundHog ($GRU$)} uses a sophisticated gate function $GRU$;
    \item {\bf GroundHog-Coverage}~\cite{Tu:2016:ACL}\footnote{\protect\url{https://github.com/tuzhaopeng/NMT-Coverage}.}:  an improved attention-based NMT model with a coverage mechanism.
\end{itemize}

\subsection{Translation Quality}
\label{sec-results}

\begin{table*}[t]
\centering
\begin{tabular}{c|ccc|ccc}
    \multirow{ 3}{*}{}   &	\multicolumn{6}{c}{GroundHog~~~~vs.~~~~GroundHog+Context Gate}\\
    \cline{2-7}
    				&	\multicolumn{3}{c|}{Adequacy}			&	\multicolumn{3}{c}{Fluency}\\
    \cline{2-7}
    				&	$<$		&	$=$		&	$>$		&	$<$		&	$=$		&	$>$\\
    \hline
    evaluator1		&	30.0\%	&	54.0\%	&	16.0\%	&	28.5\%	&	48.5\%	&	23.0\%\\
    evaluator2		&	30.0\%	&	50.0\%	&	20.0\%	&	29.5\%	&	54.5\%	&	16.0\%\\

\end{tabular}
\caption{Subjective evaluation of translation adequacy and fluency.}
\label{table-subjective-evaluation}
\end{table*}

Table~\ref{table-translation-results} shows the translation performances in terms of BLEU scores. We carried out experiments on multiple NMT variants. For example, ``2 + Context Gate ($both$)'' in Row 3 denotes integrating ``Context Gate ($both$)'' into the baseline in Row 2 (\ie GroundHog ($vanilla$)).
For baselines, we found that the gated unit (\ie $GRU$, Row 4) indeed surpasses its vanilla counterpart (\ie $\tanh$, Row 2), which is consistent with the results in other work~\cite{Chung:2014:arXiv}.
Clearly the proposed context gates significantly improve the translation quality in all cases, although there are still considerable differences among the variants:

\paragraph{Parameters}
Context gates introduce a few new parameters. The newly introduced parameters include $W_z \in \mathbb{R}^{n \times m}$, $U_z \in \mathbb{R}^{n \times n}$, $C_z \in \mathbb{R}^{n \times n'}$ in Equation~\ref{eqn-context-gate}. In this work, the dimensionality of the decoding state is $n=1000$, the dimensionality of the word embedding is $m=620$, and the dimensionality of context representation is $n'=2000$.
The context gates only introduce 3.6M additional parameters, which is quite small compared to the number of parameters in the existing models (\eg 84.3M in the ``GroundHog ($GRU$)'').

\paragraph{Over GroundHog ({\em vanilla})} We first carried out experiments on a simple decoder without gating function (Rows 2 and 3), to better estimate the impact of context gates. As shown in Table~\ref{table-translation-results}, the proposed context gate significantly improved translation performance by 4.2 BLEU points on average. It is worth emphasizing that context gate even outperforms a more sophisticated gating function (\ie GRU in Row  4). This is very encouraging, since our model only has a single gate with half of the parameters (\ie 3.6M versus 7.2M) and less computations (\ie half the matrix computations to update the decoding state\footnote{We only need to calculate the context gate once via Equation~\ref{eqn-context-gate} and then apply it when updating the decoding state. In contrast, GRU requires the calculation of an update gate, a reset gate, a proposed updated decoding state and an interpolation between the previous state and the proposed state. Please refer to~\cite{Cho:2014:EMNLP} for more details.}).

\paragraph{Over GroundHog ({\em GRU})} We then investigated the effect of the context gates on a standard NMT with GRU as the decoding activation function (Rows 4-7).  Several observations can be made.
First, context gates also boost performance beyond the GRU in all cases, demonstrating our claim that context gates are complementary to the reset and update gates in GRU.
Second, jointly controlling the information from both translation contexts consistently outperforms its single-side counterparts, indicating that a direct interaction between input signals from the source and target contexts is useful for NMT models.

\paragraph{Over GroundHog-Coverage ({\em GRU})} We finally tested on a stronger baseline, which employs a coverage mechanism to indicate whether or not a source word has already been translated~\cite{Tu:2016:ACL}. Our context gate still achieves a significant improvement of 1.6 BLEU points on average, reconfirming our claim that the context gate is complementary to the improved attention model that produces a better source context representation. Finally, our best model (Row 7) outperforms the SMT baseline system using the same data (Row 1) by 3.3 BLEU points.

\vspace{5pt}
\noindent From here on, we refer to ``GroundHog'' for ``GroundHog ($GRU$)'', and ``Context Gate'' for ``Context Gate ($both$)'' if not otherwise stated.

\paragraph{Subjective Evaluation} We also conducted a subjective evaluation of the benefit of incorporating context gates. Two human evaluators were asked to compare the translations of 200 source sentences randomly sampled from the test sets without knowing which system produced each translation. Table~\ref{table-subjective-evaluation} shows the results of subjective evaluation.
The two human evaluators made similar judgments: 
in adequacy, around 30\% of GroundHog translations are worse, 52\% are equal, and 18\% are better; while in fluency, around 29\% are worse, 52\% are equal, and 19\% are better.

\subsection{Alignment Quality}
\label{sec-alignment-quality}

\begin{table}[t]
\centering
\begin{tabular}{l|cc}
    {\bf System}					&	SAER &  AER\\
    \hline
    GroundHog				&	67.00	&	54.67\\
    ~~~+ Context Gate                    &	67.43	&	55.52\\
    \hdashline
    GroundHog-Coverage~~~                   &	64.25	&	50.50\\
    ~~~+ Context Gate				&	63.80	&	49.40\\
\end{tabular}
\caption{Evaluation of alignment quality. The lower the score, the better the alignment quality.}
\label{table-alignment-results}
\end{table}

\begin{figure*}[t]
\begin{center}
        \subfloat[GroundHog-Coverage ({\small SAER=50.80})]{
            \includegraphics[width=0.45\textwidth]{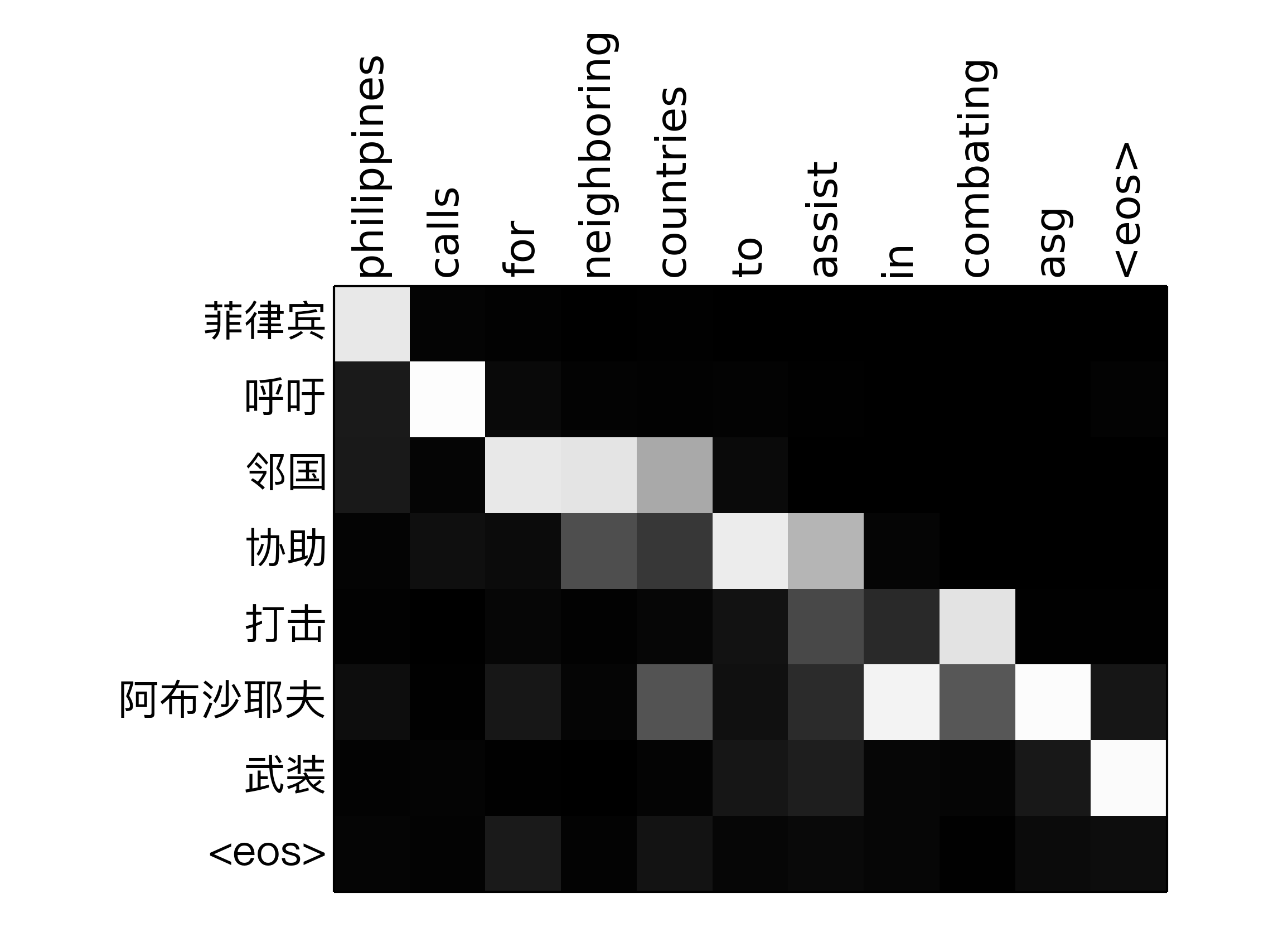}} 
        \subfloat[ + Context Gate ({\small SAER=47.35})]{
            \includegraphics[width=0.45\textwidth]{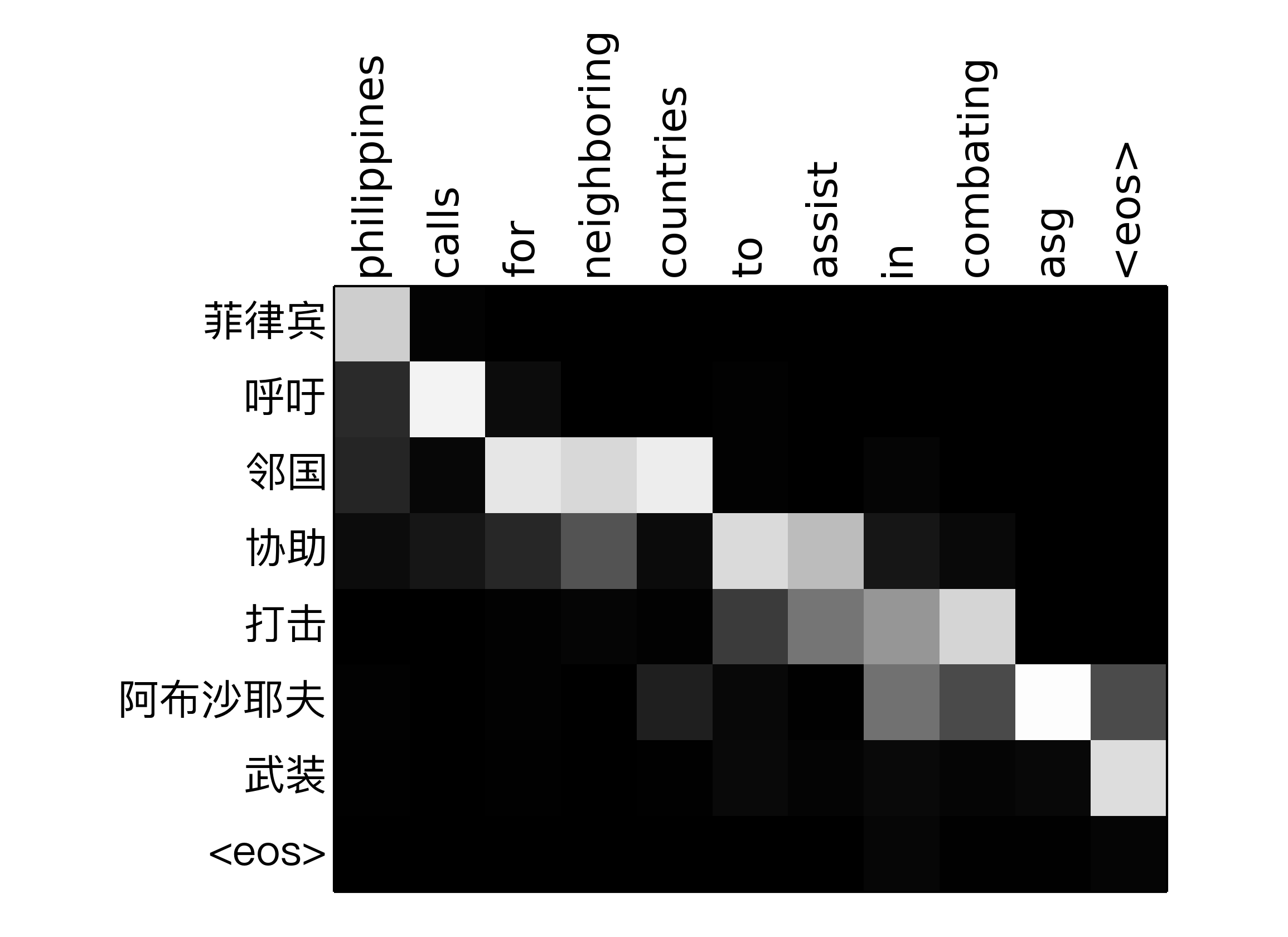}}
\end{center}
\caption{Example alignments. Incorporating context gate produces more concentrated alignments.}
\label{figure-alignment-examples}
\end{figure*}

Table~\ref{table-alignment-results} lists the alignment performances. Following~\newcite{Tu:2016:ACL}, we used the alignment error rate (AER)~\cite{Och:2003} and its variant SAER to measure the alignment quality:
\begin{eqnarray}
SAER=1-\frac{|M_{A} \times M_{S}|+|M_{A} \times M_{P}|}{|M_{A}|+|M_{S}|} \nonumber
\end{eqnarray}
where $A$ is a candidate alignment, and  $S$ and $P$ are the sets of sure and possible links in the reference alignment respectively ($S\subseteq P$).
$M$ denotes the alignment matrix, and for both $M_{S}$ and $M_{P}$ we assign the elements that correspond to the existing links in $S$ and $P$ probability $1$ and the other elements probability $0$. 
In this way, we are able to better evaluate the quality of the soft alignments produced by attention-based NMT.

We find that context gates do not improve alignment quality when used alone. When combined with coverage mechanism, however, it produces better alignments, especially one-to-one alignments by selecting the source word with the highest alignment probability per target word (i.e., AER score).
One possible reason is that better estimated decoding states (from the context gate) and coverage information help to produce more concentrated alignments, as shown in Figure~\ref{figure-alignment-examples}.

\begin{table*}[t]
\centering
\begin{tabular}{c|l|l|llll}
    \#	&	{\bf System}	&	{\bf Gate Inputs}	&	{\bf MT05}  &  {\bf MT06}	  &	{\bf MT08}  &  {\bf Ave.}\\
    \hline
    1	&	GroundHog							&	--					&	30.61		&	31.12		&	23.23	&	28.32\\
    \hdashline
    2	&	1 + Gating Scalar						&	$t_{i-1}$				&	31.62$^*$	&	31.48		&	23.85	&	28.98\\
    \hdashline
    3	&	\multirow{ 1}{*}{1 + Context Gate ($source$)}  	&	$t_{i-1}$				&	31.69$^*$	&	31.63		&	24.25$^*$	&	29.19\\
   \hdashline
   4	&	\multirow{ 3}{*}{1 + Context Gate ($both$)}  	&	$t_{i-1}$				&	32.15$^*$	&	32.05$^*$	&	24.39$^*$	&	29.53\\
   5	&	 									&	$t_{i-1}$, $s_i$			&	31.81$^*$	&	32.75$^*$	&	25.66$^*$	&	30.07\\
   6	&	 									&	$t_{i-1}$, $s_i$, $y_{i-1}$	&	33.52$^*$	&	33.46$^*$	&	24.85$^*$	&	30.61\\
\end{tabular}
\caption{Analysis of the model architectures measured in BLEU scores. ``Gating Scalar'' denotes the model proposed by (Xu et al.,2015) in the image caption generation task, which looks at only the previous decoding state $t_{i-1}$ and scales the whole source context $s_i$ at the vector-level. To investigate the effect of each component, we list the results of context gate variants with different inputs (\eg the previously generated word $y_{i-1}$). ``*'' indicates statistically significant difference ($p < 0.01$) from ``GroundHog''.}
\label{table-component-analysis}
\end{table*}

\subsection{Architecture Analysis}
\label{sec-component-analysis}

Table~\ref{table-component-analysis} shows a detailed analysis of architecture components measured in BLEU scores. Several observations can be made:

\begin{itemize}
\item {\bf Operation Granularity} (Rows 2 and 3): Element-wise multiplication (\ie Context Gate ($source$)) outperforms the vector-level scalar (\ie Gating Scalar), indicating that precise control of each element in the context vector boosts translation performance.
\item {\bf Gate Strategy} (Rows 3 and 4): When only fed with the previous decoding state $t_{i-1}$, Context Gate ($both$) consistently outperforms Context Gate ($source$), showing that jointly controlling information from both source and target sides is important for judging the importance of the contexts.
\item {\bf Peephole connections} (Rows 4 and 5): Peepholes, by which the source context $s_i$ controls the gate, play an important role in the context gate, which improves the performance by 0.57 in BLEU score.
\item {\bf Previously generated word} (Rows 5 and 6): Previously generated word $y_{i-1}$ provides a more explicit signal for the gate to judge the importance of contexts, leading to a further improvement on translation performance.
\end{itemize}

\subsection{Effects on Long Sentences}

\begin{figure*}[t]
\centering
\includegraphics[width=0.38\textwidth]{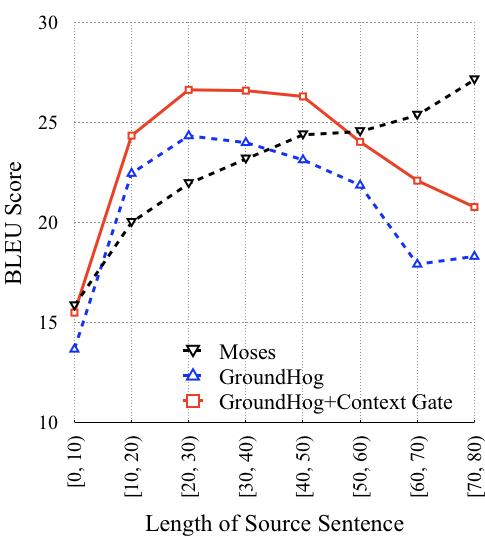} \hspace{0.05\textwidth}
\includegraphics[width=0.38\textwidth]{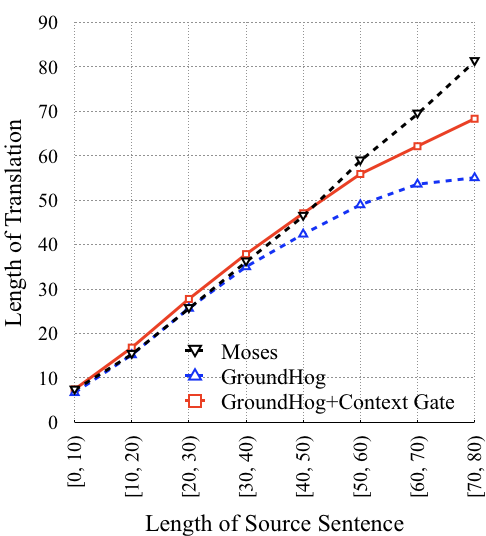}
\caption{Performance of translations on the test set with respect to the lengths of the source sentences. Context gate improves performance by alleviating in-adequate translations on long sentences.}
\label{figure-sentence-len}
\end{figure*}

We follow~\newcite{Bahdanau:2015:ICLR} and group sentences of similar lengths together. Figure~\ref{figure-sentence-len} shows the BLEU score and the averaged length of translations for each group.
GroundHog performs very well on short source sentences, but degrades on long source sentences (\ie $\geqslant30$), which may be due to the fact that source context  is not fully interpreted. Context gates can alleviate this problem by balancing the source and target contexts, and thus improve decoder performance on long sentences. In fact, incorporating context gates boost translation performance on all source sentence groups.

We confirm that context gate weight $z_i$ correlates well with translation performance. In other words, translations that contain higher $z_i$ (\ie source context contributes more than target context) at many time steps are better in translation performance.
We used the mean of the sequence $z_1, \dots, z_i, \dots, z_I$ as the gate weight of each sentence. 
We calculated the Pearson Correlation between the sentence-level gate weight and the corresponding improvement on translation performance (\ie BLEU, adequacy, and fluency scores),\footnote{We use the average of correlations on subjective evaluation metrics (\ie adequacy and fluency) by two evaluators.} as shown in Table~\ref{table-correlation}.
We observed that context gate weight is positively correlated with translation performance improvement and that the correlation is higher on long sentences.

\begin{table}[t]
\centering
\renewcommand\arraystretch{1.1}
\begin{tabular}{l|c|c|c}
    {\bf Length}	&	{\bf BLEU}	&	{\bf Adequacy}	&	{\bf Fluency}\\
    \hline
    $<30$			&	0.024		&	0.071	&	0.040\\
    $\geqslant30$	&	0.076		&	0.121	&	0.168\\
\end{tabular}
\caption{Correlation between context gate weight and improvement of translation performance. ``Length'' denotes the length of source sentence. ``BLEU'', ``Adequacy'', and ``Fluency'' denotes different metrics measuring the translation performance improvement of using context gates.}
\label{table-correlation}
\end{table}

As an example, consider this source sentence from the test set:
\begin{quote}
{\em zh{\= o}uli{\` u}  zh{\` e}ngsh{\`\i}  y{\=\i}nggu{\' o}  m{\'\i}nzh{\` o}ng  d{\` a}o  ch{\= a}osh{\`\i}  c{\v a}ig{\` o}u  de  g{\=a}of{\= e}ng  sh{\'\i}k{\` e},  d{\= a}ngsh{\'\i}  14  ji{\=a}  ch{\= a}osh{\`\i}  de  gu{\= a}nb{\`\i}  l{\`\i}ng  y{\=\i}nggu{\' o}  zh{\` e}  ji{\= a}  zu{\`\i}  d{\` a}  de  li{\' a}nsu{\v o}  ch{\= a}osh{\`\i}  s{\v u}nsh{\=\i}   sh{\` u}b{\v a}iw{\` a}n  y{\=\i}ngb{\` a}ng  de  xi{\= a}osh{\` o}u  sh{\= o}ur{\` u}  .}
\end{quote}
GroundHog translates it into:
\begin{quote}
	{\em twenty - six london supermarkets were closed at a peak hour of the british population in the same period of time .}
\end{quote}
which almost misses all the information of the source sentence.
Integrating context gates improves the translation adequacy:
\begin{quote}
	{\em this is exactly the peak days British people buying the supermarket . the closure {\bf of the 14 supermarkets of the 14 supermarkets} that the largest chain supermarket in england lost several million pounds of sales income .}
\end{quote}
Coverage mechanisms further improve the translation by rectifying over-translation (e.g., ``{\em of the 14 supermarkets}'') and under-translation (e.g., ``{\em saturday}'' and ``{\em at that time}''):
\begin{quote}
	{\em \underline{saturday} is the peak season of british people 's purchases of the supermarket . \underline{at} \underline{that time} , the closure of 14 supermarkets made the biggest supermarket of britain lose millions of pounds of sales income .}
\end{quote}

\section{Conclusion}

We find that source and target contexts in NMT are highly correlated to translation {\em adequacy} and {\em fluency}, respectively. Based on this observation, we propose using context gates in NMT to dynamically control the contributions from the source and target contexts in the generation of a target sentence, to enhance the adequacy of NMT.
By providing NMT the ability to choose the appropriate amount of information from the source and target contexts, one can alleviate many translation problems from which NMT suffers. Experimental results show that NMT with context gates achieves consistent and significant improvements in translation quality over different NMT models.

Context gates are in principle applicable to all sequence-to-sequence learning tasks in which information from the source sequence is transformed to the target sequence (corresponding to {\em adequacy}) and the target sequence is generated (corresponding to {\em fluency}). In the future, we will investigate the effectiveness of context gates to other tasks, such as dialogue and summarization.
It is also necessary to validate the effectiveness of our approach on more language pairs and other NMT architectures (\eg using LSTM as well as GRU, or multiple layers).

\section*{Acknowledgement}

This work is supported by China National 973 project 2014CB340301.
Yang Liu is supported by the National Natural Science Foundation of China (No. 61522204) and the 863 Program (2015AA015407).
We thank action editor Chris Quirk and three anonymous reviewers for their insightful comments.

\balance
\bibliography{all}
\bibliographystyle{acl2012}

\end{document}